\documentclass{article}

\usepackage{microtype}
\usepackage{graphicx}
\usepackage{subfigure}
\usepackage{booktabs} 
\usepackage{multirow}

\usepackage{hyperref}
\usepackage{placeins}

\usepackage[accepted]{aux/icml2024}

\usepackage{amsmath}
\usepackage{amssymb}
\usepackage{mathtools}
\usepackage{amsthm}

\usepackage[capitalize,noabbrev]{cleveref}

\newcommand{\icmlInternship}{\textsuperscript{*}Work done during an internship with AWS AI Labs}

\theoremstyle{plain}
\newtheorem{theorem}{Theorem}[section]
\newtheorem{proposition}[theorem]{Proposition}

\theoremstyle{definition}

\theoremstyle{remark}

\newcommand{\A}{A}
\newcommand{\B}{B}
\newcommand{\C}{C}

\newcommand{\argmax}{\operatorname*{arg\,max}}
\newcommand{\argmin}{\operatorname*{arg\,min}}

\usepackage[textsize=tiny]{todonotes}

\icmltitlerunning{RAG as In-Context Optimization}

\newcommand{\methodname}{RICO }

\begin{document}

\twocolumn[
\icmltitle{
Maximally-Informative Retrieval for State Space Model Generation 
}
]

\icmlsetsymbol{intern}{*}

\begin{icmlauthorlist}
\icmlauthor{Evan Becker}{intern,ucla,amazon}
\icmlauthor{Benjamin Bowman}{amazon}
\icmlauthor{Matthew Trager}{amazon}
\icmlauthor{Tian Yu Liu}{intern,ucla,amazon}
\icmlauthor{Luca Zancato}{amazon}
\icmlauthor{Wei Xia}{amazon}
\icmlauthor{Stefano Soatto}{amazon}
\end{icmlauthorlist}

\icmlaffiliation{ucla}{Department of Computer Science, UCLA}
\icmlaffiliation{amazon}{AWS AI Labs}
\icmlcorrespondingauthor{Evan Becker}{evbecker@amazon.com}

\icmlkeywords{Machine Learning, ICML}

\vskip 0.3in


\printAffiliationsAndNotice{\icmlInternship} 

\begin{abstract}
Given a query and dataset, the optimal way of answering the query is to make use all the information available. Modern LLMs exhibit impressive ability to memorize training data, but data not deemed important during training is forgotten, and information outside that training set cannot be made use of. 
Processing an entire dataset at inference time is infeasible due to the bounded nature of model resources (e.g.\ context size in transformers or states in state space models), meaning we must resort to external memory. 
This constraint naturally leads to the following problem: How can we decide based on the present query and model, what among a virtually unbounded set of known data matters for inference?
To minimize model uncertainty for a particular query at test-time, we introduce Retrieval In-Context Optimization (\methodname\!), a retrieval method that uses gradients from the LLM itself to learn the optimal mixture of documents for answer generation. 
 Unlike traditional retrieval-augmented generation (RAG), which relies on external heuristics for document retrieval, our approach leverages direct feedback from the model.
Theoretically, we show that standard top-$k$ retrieval with model gradients can approximate our optimization procedure, and provide connections to the leave-one-out loss.
We demonstrate empirically that by  minimizing an unsupervised loss objective in the form of question perplexity, we can achieve comparable retriever metric performance to BM25 with \emph{no finetuning}. Furthermore, when evaluated on quality of the final prediction, our method often outperforms fine-tuned dense retrievers such as E5. 
\end{abstract}


\section{Introduction}

A key advantage of large language models (LLMs) is their in-context learning (ICL) capability, which allows them to improve predictions by adapting to examples in the input \cite{dong2022survey}.
Retrieval-Augmented Generation (RAG) takes advantage of ICL by trying to inject the most relevant contexts into the prompt before the query \cite{gao2023retrieval,fan2024survey}. This procedure is more efficient and oftentimes more accurate than finetuning the language model separately for each application \cite{ovadia2023fine}. Furthermore, the data store used for context retrieval can be updated on the fly as new documents are added (or outdated ones removed). 
Despite the potential benefits, a complex and dynamically evolving set of documents leads to the following challenge: which subset of contexts should be considered relevant for a given model and query?

Current retrieval systems do not always find the contexts that improve accuracy on question-answering tasks. Authors in \cite{li2024retrieval} explore failure modes in RAG and find common instances of where the query is too complex for the retriever to understand, such as in multi-step reasoning or when the question is implicit. Another recent work \cite{jin2024long} observes `inverted-U' shaped accuracy curves where performance declines after a certain number of documents are added to the prompt, even for state-of-the-art retrievers. We contend that these issues occur due to the majority of existing retrievers operating in an `open-loop' configuration. That is, they respond with the same documents for the same query regardless of whether the LLM actually makes use of the information. Furthermore, most retrievers rank documents based on \emph{similarity} to the query, not their \emph{informativeness}. Both bag-of-words based methods such as BM25 \citep{roberston2009probabilistic} and semantic models trained with a fixed set of `ground truth' query, document pairs can suffer from this objective misalignment.

Determining whether additional context benefits an LLM's generation (model-aware retrieval) has therefore become an important line of research (see \cref{sec:related_model_aware}). Yet most methods here rely on expensive retriever fine-tuning that must be performed for each new task, and repeated with every model upgrade.

\begin{figure*}[h]
    \centering
    \includegraphics[width=0.95\textwidth]{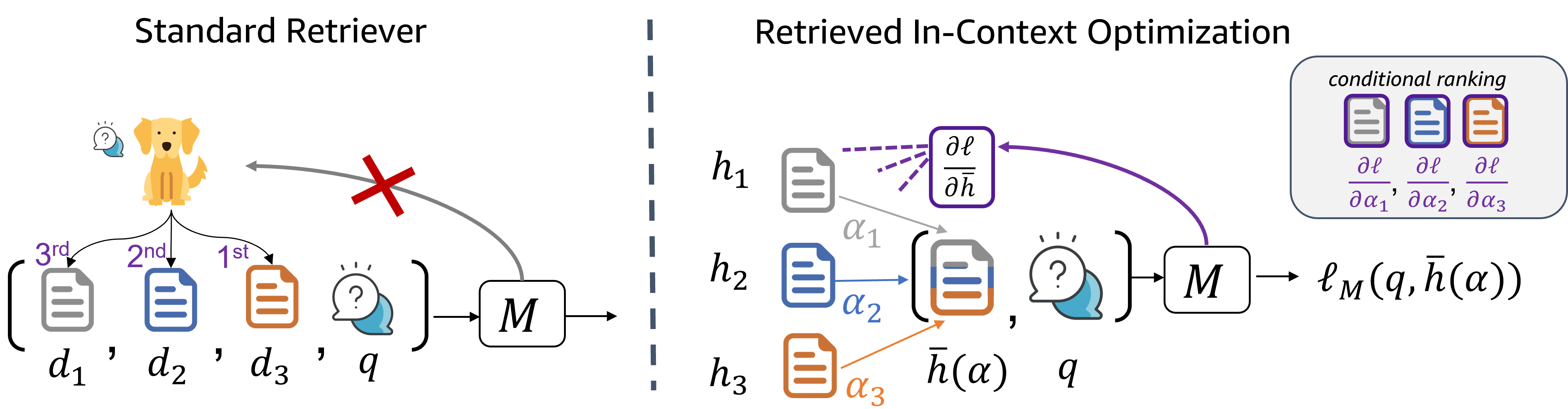}
    \caption{Comparison of `open loop' retrieval in standard RAG systems (left) with our context optimization method using model gradients (right). This method allows us to `learn' useful document mixtures at test time.}
    \label{fig:enter-label}
\end{figure*}

\subsection{Contributions}

We propose a new framework for model-aware context retrieval and optimization that works using gradients from state space models (SSMs) with no finetuning.

\begin{itemize}
    \item In \cref{sec:method} we take advantage of a key decomposition property of SSMs to pose a continuous relaxation of the context reranking problem. 
    \item We show that this optimization procedure is approximated by taking the inner product of model gradients and precomputed document states.
    \item Empirically, we find that reranking using our method produces better retrieval-side metrics than BM25, and when evaluated on quality of the final generation, often outperforms standard semantic retrievers such as E5. This advantage becomes even greater in long-context scenarios.
\end{itemize}

\section{Related Work}

\subsection{Model-Aware Retrieval}\label{sec:related_model_aware}
A variety of methods have been recently introduced to incorporate some feedback from the generating model into the retrieval and prompt augmentation steps. These methods range in terms of complexity from heuristically sampling model outputs to requiring full training of the generating model with a custom dataset.

\paragraph{Heuristic Approaches}
While standard retrievers examine document similarity with the user query, methods such as (HyDE) \cite{gao2022precise} and (ReFEED) \cite{yu2023improving} first ask a model to generate initial documents or draft answers, which are then used to query the document store. The IRCoT method \citep{trivedi2022interleaving} takes this idea one step further by retrieving documents at each step of a chain-of-thought (CoT) generation process. Authors in \cite{wang2023self} explore methods such as direct prompting and KNN retrieval of similar questions for detecting when the generating model needs external information. Note that the most succesful proposed method (KNN) relied on a large dataset of previously answered questions with ground truth evaluation.

\paragraph{Retriever Finetuning}
Retrievers can also be finetuned on data from a specific generating model before deployment.
For example, REPLUG freezes the generating LM and trains a dense retriever to return the top documents that improve model accuracy \cite{shi2023replug}. However this procedure is expensive as it requires the entire datastore to be re-indexed multiple times throughout training. Alternatively authors in  \cite{xu2023recomp} propose RECOMP, which performs prompt augmentation by training a separate compressive model to summarize retrieved documents in a way that the generating LM can better utilize. Lastly, authors in \cite{rubin2021learning} train a retriever to grab similar prompt + answer combos for the purposes of in-context learning (ICL). 

\paragraph{Custom Generating Models}
Instead of altering the retrieving model, the generating model itself can be trained to interact with databases and search engines. Models like Toolformer \cite{schick2024toolformer}, which are trained to use API calls to external websites such as Wikipedia, can potentially be used for RAG applications. Self-RAG, proposed in \cite{asai2023self}, specifically trains the generating model to produce new tokens that signal when the model needs external information, and whether that information is useful. The training dataset relies on an expert model (GPT-4) to annotate passages. Repoformer \cite{wu2024repoformer}, is a code-specific generating model that is also trained to predict when its predictions could benefit from seeing additional context files. Their dataset does not require an expert model but does compare whether model predictions improve before and after seeing sections of code. The main drawback of this approach is the computational expense of creating the training dataset. It is also not certain the model predictions will improve if document sets are altered.

\paragraph{Token-level Methods}
Perhaps most similar to our proposal is a line of work that uses token-level uncertainty of the generating model to determine retrieval.
In \cite{jiang2023active}, authors propose an active retrieval method (FLARE) that retrieves documents whenever token-level probabilities fall below a certain threshold.
DRAGIN builds upon this method by using the self-attention weights of the uncertain token to find the most impactful token keywords to search for \cite{su2024dragin}. \citet{yao2024seakr} propose SEAKR, which uses an uncertainty estimator on sampled token hidden states to rerank documents according to what reduces their uncertainty metric the most.
While we will also use token level metrics from the generating model as a retrieval signal, our method provides a principled way of retrieving and ranking the documents as opposed to a more heuristic keyword search. Our method is also more computationally efficient than uncertainty scoring methods like SEAKR which require multiple passes through the generating model for each individual document.

\subsection{Language Models as Text Embedders}
In parallel to the exploration of RAG, another line of work has attempted to convert LLMs into efficient general text embedders. These embeddings are evaluated on a range of downstream tasks that include retrieval and reranking and achieve near state of the art on these benchmarks \cite{muennighoff2022mteb}. Embedding models such as e5-7b--mistral \cite{wang2023improving} and bge-en-icl \cite{li2024making} finetune the last [EOS] token embedding for decoding models, while NV-Embed \cite{lee2024nv} and LLM2vec \cite{behnamghader2024llm2vec} finetune after removing the causal mask and using mean pooling across token positions. GritLM proposes a dual method that maintains the causal mask during generation while removing it when in embedding mode \cite{muennighoff2024generative}. However, despite their differences all methods rely on finetuning, resulting in additional costs for each new model adaptation.

\subsection{Gradients as Features}
Previous works have explored different methods of utilizing gradients as features for applications such as task representation \cite{achille2019task2vec}, transfer learning \cite{mu2020gradients}, and anomaly detection \cite{kwon2020backpropagated}. To our knowledge this work is the first to utilize gradients specifically for document retrieval.

\section{Method}\label{sec:method}
\subsection{Notation}
For any sequence of matrices $M_0,\dots,M_N$, we denote their matrix product as $M_{0:N}:=\prod_{i=1}^N M_i$. We will denote our auto-regressive language model as $\mathcal{M}$. We will also represent the concatenation of tokens (and token sequences) as $(x_1,x_2)$, where tokens in $x_1$ appear before $x_2$.

\subsection{Separability of Model Dynamics}
We will first introduce our method through the perspective of state space models (SSMs) such as Mamba \cite{gu2023mamba}. In \cref{sec:transformers} we also show that our method applies to a larger class of models including linear attention transformers \cite{yang2023gated,dao2024transformers,zancato2024b}. We start with the time varying dynamics using input $x_t \in \mathbb{R}^m$, output $y_t\in \mathbb{R}^p$, and hidden state $h_t \in \mathbb{R}^n$
\begin{align}
    h_{t} &= \A_t h_{t-1} + \B_t x_t \nonumber \\
    y_t &= \C_t h_t
\end{align}
Where $\A_t\in \mathbb{R}^{(n,n)}$, $\B_t\in \mathbb{R}^{(n,m)}$, $\C_t\in \mathbb{R}^{(p,n)}$ are dependent on the current input. It is standard practice in control theory to unroll the recurrent dynamics into the following matrix product between the controllability matrix and input sequence $X:=[x_0^\intercal \: \dots \: x_t^\intercal]^\intercal$. 
\begin{align}
    y_t =  \C_t \begin{bmatrix}
         \A_{t:1}\B_0 & \A_{t:2} \B_1 & \dots & \A_t\B_{t-1}& \B_t
        \end{bmatrix} X
\end{align}
Next, assume our input sequence can be partitioned into two chunks, which we will denote the context $X_c := [x_0^\intercal\dots x_\tau^\intercal]^\intercal$ and the query $X_q := [x_{\tau+1}^\intercal\dots x_{t}^\intercal]^\intercal$, for any $\tau$ such that $0\leq\tau\leq t$. 
Note that for output $y_t$, we can decompose our map as
\begin{align}\label{eq:separable}
    y_t =& \C_t \begin{bmatrix}
         \A_{t:1}\B_0 & \A_{t:2} \B_1 & \dots & \A_{t:\tau+1}\B_{\tau}
        \end{bmatrix} X_c  \nonumber\\ 
        &+ \C_t \begin{bmatrix}
         \A_{t:\tau+2}\B_{\tau+1} & \A_{t:\tau+3} \B_{\tau+2} & \dots &\B_{t}
        \end{bmatrix} X_{q} \nonumber \\
        =& \C_t\A_{t:\tau+1}h_c+\C_t h_{q},
\end{align}
Here $h_c$ is the hidden state after processing the inputs in the context, and $h_{q}$ is the state after separately processing inputs from the query. This separability means that for any fixed query sequence, the output depends on the context sequence only through the hidden state $h_c$ (i.e. $y_t\leftarrow h_c \leftarrow X_c$). We can use \cref{eq:separable} to further derive the function for all outputs associated with the query $Y_q:=[y_{\tau+1}^\intercal,\dots,y_t^\intercal]^\intercal$:
\begin{align} \label{eq:separable_vec}
    Y_q &= \mathbf{H}_q h_c + \mathbf{M}_q X_q, 
\end{align}
where $\mathbf{H}_q$ is defined as
\begin{align}
        \mathbf{H}_q&:=\begin{bmatrix} \C_{\tau+1}\A_{\tau+1}\\
    \C_{\tau+2}\A_{\tau+2:\tau+1} \\ 
    \vdots\\
    \C_{t}\A_{t:\tau+1}\end{bmatrix},
\end{align}
and $\mathbf{M}_q$ is the block lower triangular matrix
\begin{align}
        \mathbf{M}_q&:=\begin{bmatrix} \C_{\tau+1}\B_{\tau+1} & 0  &\dots &0\\
    \C_{\tau+2}\A_{\tau+2}\B_{\tau+1} &\C_{\tau+2}\B_{\tau+2}  & \dots &0\\ 
    \vdots&\vdots &\ddots &\vdots\\
    \C_{t}\A_{t:\tau+2}\B_{\tau+1} & \C_{t}\A_{t:\tau+3} \B_{\tau+2}  & \dots &\C_{t}\B_{t} \end{bmatrix}.
\end{align}
Again, we can see that the output is still a linear function of the context hidden state, as the matrix $\mathbf{H}_q$ is constant for any fixed query. 

\subsection{Continuous Relaxation of the RAG objective}
Given some set of documents $\mathcal{D}:=\{d_1,\dots,d_N\}$, each consisting of a sequence of tokens in a fixed alphabet $\mathcal{A}$, and a query, answer pair $(q,a)$ (also token sequences), we would like to find the best ordering of retrieved documents $\mathcal{D}^*_q$ out of all permutations of subsets of $k$ documents $\mathcal{P}_k(\mathcal{D})$ that maximizes the likelihood of the answer under the model. 
\begin{align}
    \mathcal{D}^*_q = \argmax_{(d_1\dots d_k)\in\mathcal{P}_k(\mathcal{D})} P_{\mathcal{M}}(a|d_1,\dots,d_k,q)
\end{align}
Without any further assumptions on our probability function and inputs, this is a combinatorial optimization problem, as there are $\frac{N!}{(N-k)!}$ possible document orderings to evaluate. Another practical issue with this formulation is that we typically do not have access to the ground truth answer during retrieval. Therefore to tackle our retrieval problem, we will need both a reliable proxy for answer likelihood and a way of relaxing the combinatorial search space.

\paragraph{Question Likelihood as a Proxy:}
In \cite{muennighoff2022sgpt}, authors examine the cross-encoder properties of GPT models and demonstrate that using the sum of token log probabilities in the query $\sum_{i=0}^T \log P_\mathcal{M}(q_i|d)$ can serve as a good scoring function for each document . One drawback to the original method is that each pairwise score requires reprocessing both pieces of text (document and query) together, leading to a quadratic number of calls to the model. We will show how our method can alleviate this issue (requiring only a linear number of calls) when combining this loss function with properties of state space models.   

\paragraph{State Space Relaxation:} Consider for now a restricted version of our problem setting for a single SSM layer output:  given inputs from $k$ documents, we aim to find the optimal ordering such that our output is close to some target $y_t^*$. 
We know that for any concatenation of documents $(d_1\dots d_k)$ where inputs from document $d_i$ appear in the span $(\tau_{i-1},\tau_i]$ and inputs from the query in the span $(\tau_k,T]$, we can use \cref{eq:separable} to write outputs $y_t$ as 
\begin{align}
    y_t=& \C_t(\A_{t:\tau_1+1} h_{d_1}+\A_{t:\tau_2+1} h_{d_2}\dots+\A_{t:\tau_k+1} h_{d_k}+h_{q}),
\end{align}
where $h_{d_i}$ is the hidden state for $d_i$ processed independently.
Previous work has observed that changing the ordering of these documents corresponds to changing $\A_{t:\tau}$ matrices, which depend on inputs from other documents \cite{liu2025picaso}. In the case of Mamba2, these matrices are scalar multiples of the identity matrix leading to the simplification
\begin{align}
    y_t=& \C_t(\sum_{i=1}^k\alpha_{t:\tau_i+1} h_{d_i}+h_{q}).
\end{align}
The key observation in this work is that we can turn our combinatorial optimization problem into a continuous one for some loss function $\mathcal{L} = \ell(y_t^*-y_t) \in [0,1]$ by relaxing the constraints on these scalars:
\begin{align} \label{eq:continuous_relaxation}
    \hat{\mathbf{\alpha}} &= \argmin_{\alpha_1,\dots \alpha_k \in [0,1]} \ell \left(y_t^* - \C_t(\sum_{i=1}^k\alpha_{i} h_{d_i}+h_{q})\right) 
\end{align}
This can be interpreted as finding the best linear combination of document states $\bar{h}(\alpha) = \sum_{i=1}^k\hat{\alpha}_{i} h_{d_i}$ that minimizes our error for a specific query input. 

Note that in reality our loss $\mathcal{L}$ will be a function of token logits rather than state outputs, and we will be working over multiple layers of SSM blocks. However we show empirically that this continuous relaxation still performs well. 

\subsection{Gradients for Top-$k$ Retrieval}
We use the following simple observation to motivate our use of gradients in top-$k$ retrieval. Suppose we initialize the weights of our combined document state as uniform over the data store such that $\bar{h}(\bar{\alpha})= \sum_{i=1}^N\frac{1}{N} h_{d_i}$. If we want to perform one step of gradient descent on the loss from \cref{eq:continuous_relaxation} with learning rate $\eta$ to find the updated document weight $\alpha_i'$, we can apply the chain rule as follows:

\begin{align}
    \alpha_i' = \alpha_i - \eta \frac{\partial \mathcal{L}}{\partial\alpha_i} = \alpha_i - \eta h_{d_i}^\intercal \frac{\partial \mathcal{L}}{\partial\bar{h}}
\end{align}

Therefore to find the $k$ largest $\alpha_i'$ after uniform initialization, we need only to find the $k$ largest inner products between document states and (negative) loss gradients w.r.t. the input $\langle h_{d_i},-\frac{\partial \mathcal{L}}{\partial\bar{h}}\rangle$. Document states can be precomputed independently of the query, meaning top-$k$ retrieval over this state index effectively approximates one step of gradient descent on our loss. 

\begin{figure}
    \centering
    \includegraphics[width=\linewidth]{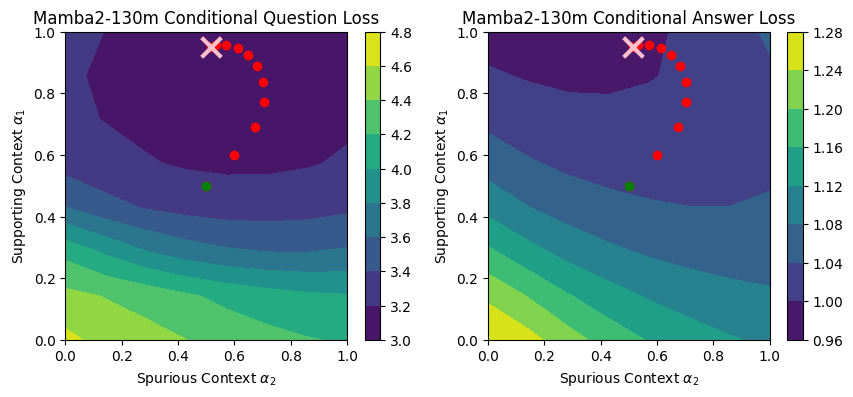}
    \caption{Cross-entropy loss landscapes of the Mamba2-130m model for token sequences conditioned on two different documents from the HotpotQA dataset. Darker colors indicate that the token sequence is evaluated as more likely. The loss over the token sequence is computed over an example question (left) and associated answer (right). Y-axis represents the document weight of the ground truth supporting document while the X-axis is the magnitude of the spurious document.  Red dots indicate the trajectory of a gradient descent algorithm (AdamW with $\eta=1e^{-1}$) on the question loss landscape starting from uniform initialization (green dot). }
    \label{fig:viz_loss}
\end{figure}

Furthermore, in the case where $\ell(y_t^*-y_t)$ is convex (e.g. an $\ell_2$ norm), we have the following simple guarantee for these inner products. While this is not the exact case in practice, we illustrate in \cref{fig:viz_loss} (and further in \cref{sec:loss_landscape}) that the marginal loss in the document directions are often convex, which provides tentative evidence that the landscape may be approximately convex.

\begin{proposition}
    Suppose our loss function is convex in the direction of $h_{d_i}$. Then the retrieval score produced by the inner product $\langle h_{d_i},\frac{\partial \mathcal{L}}{\partial\bar{h}}\rangle$ is an upper bound on the (order agnostic) leave-one-out loss.
    \begin{proof}
        With slight abuse of notation we will denote $\mathcal{L}(\alpha)$ as our loss as a function of document weights. Define the order agnostic leave-one-out loss as 
        $\hat{loo}(d_i) :=\mathcal{L}(\bar{\alpha})-\mathcal{L}(\bar{\alpha}_{\setminus i})$,
        where $\bar{\alpha} = (\frac{1}{N},\dots,\frac{1}{N})^\intercal$ and $\bar{\alpha}_{\setminus i} = (\frac{1}{N},\dots,0, \dots\frac{1}{N})^\intercal$, i.e. $\hat{loo}(d_i)$ will be positive if removing document $d_i$ lowers the loss. Then using Jensen's inequality we have
        \begin{align}
            \hat{loo}(d_i) &\leq \nabla_\alpha\mathcal{L}(\bar{\alpha})^\intercal(\bar{\alpha}-\bar{\alpha}_{\setminus i}) \nonumber \\
            &= \frac{1}{N} h_{d_i}^\intercal\frac{\partial\mathcal{L}(\bar{\alpha})}{\partial h}\leq h_{d_i}^\intercal\frac{\partial\mathcal{L}(\bar{\alpha})}{\partial h}
        \end{align}
    \end{proof}
\end{proposition}

We can likewise interpret gradients $-\frac{\partial \ell}{\partial\alpha_i}$ for non-uniform document weights  as \emph{conditional document rankings}. That is, given the current information available to the model in the form of $\bar{h}(\alpha)$, which documents will best minimize the remaining model loss. 

\paragraph{Warm Start Initialization} Note that for larger document stores using a uniform prior for document weights may be far from the optimal $\alpha^*$. As the true loss function we are using is non-convex it may make sense to initialize the document weights using a coarse retriever such as BM25, ideally putting us in a locally convex region around the local minima.


\section{Experiments}
In this section we evaluate our method empirically on four standard retrieval and question answering datasets: MS MARCO \citep{bajaj2016ms}, HotpotQA \citep{yang2018hotpotqa}, MuSiQue \citep{trivedi2022musique}, and 2WikiMultihopQA \citep{ho2020constructing}. These datasets each provide a set of ground-truth documents (typically 10-20) for each question as well as ground-truth answers. 
For a longer context retrieval task we make use of TriviaQA, which is a reading comprehension dataset that provides full articles for reference. We provide more detailed statistics for the dataset samples used as well as illustrative examples in \cref{sec:dataset_details}.

The default version of \methodname which we will evaluate uses a few steps of a gradient descent procedure (AdamW) to learn document weights over a fixed set of documents after uniform initialization. This `reranking' of a smaller document set can be seen as similar to using warm start initialization over a larger document store. In \cref{sec:warm_start_compression}, we demonstrate that using a warm start initialization over a larger document store produces similar relative performance compared to baselines, therefore we primarily focus on this default method for computational efficiency purposes (see \cref{sec:discussion} for discussion).
For some experiments we also compare using a single step of gradient retrieval, and also when computing the gradient without initially conditioning on any documents (equivalent to starting with all $\alpha_i=0$). 

\begin{figure}
    \centering
    \includegraphics[width=\linewidth]{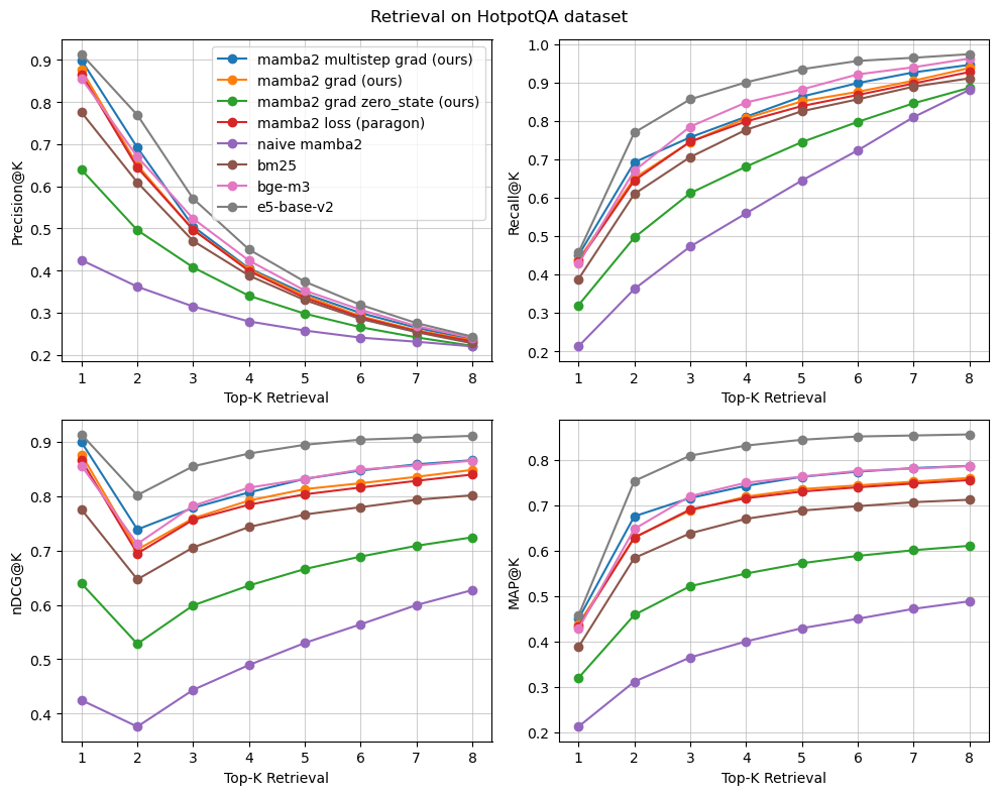}
    \caption{Performance of different retrieval methods on a subset of the HotpotQA dataset (1k samples), using four different metrics: precision (top left), recall (top right), normalized discounted cummulative gain (bottom left), and mean average precision (bottom right). The hyperparameter $k$ (x-axis) illustrates how metrics change as the number of documents retrieved increases. We compare three variations of \methodname using gradients from Mamba2-130m to baselines, as well as to a naive cosine similarity between document and query states (naive mamba2), and to using the loss score for each invidiual document directly (mamba2 loss paragon).  }
    \label{fig:hotpot_curves}
\end{figure}

\subsection{Retrieval Metrics}
\begin{table*}[!htbp]
    \centering
    \resizebox{0.9\textwidth}{!}{
    \begin{tabular}{llrrrrr}
    \toprule
    \bfseries Initial &\bfseries Reranker & MS MARCO (↑) & HotpotQA (↑) & MuSiQue (↑)& 2WikiMultihopQA (↑) & Average\\
    \midrule
    \multirow{6}{*}{Ground Truth} & BM25& 0.560 & 0.698 & 0.385 & 0.817 & 0.615\\
    &BGE-M3 & 0.668* & 0.723* & 0.639 & 0.839 & 0.717*\\
    &E5-base & 0.739* &  0.762 & 0.450 &  0.908 & 0.715*\\
    &Mamba2-130m \methodname (ours)& 0.539 & 0.713 & 0.444 & 0.852 & 0.637\\
    &Mamba2-1.3b \methodname (ours)& 0.517 & 0.703 & 0.448 & 0.847 & 0.629\\
    &Mamba2-2.7b \methodname (ours)& 0.516 & 0.696 & 0.437 & 0.839 & 0.622\\
    \midrule
    \multirow{6}{*}{BM25} &BM25 & 0.438 & 0.676 & 0.378 & 0.590 & 0.520\\
    &BGE-M3 & 0.592* & 0.720* & 0.441 & 0.671 & 0.606*\\
    &E5-base & 0.639* & 0.759 & 0.468 & 0.719 & 0.646*\\
    &Mamba2-130m \methodname (ours)& 0.442 & 0.711 & 0.458 & 0.671 & 0.57\\
    &Mamba2-1.3b \methodname (ours)& 0.422 & 0.696 & 0.440 & 0.647 & 0.551\\
    &Mamba2-2.7b \methodname (ours)& 0.410 & 0.692 & 0.427 & 0.641 & 0.543\\
    \bottomrule
    \end{tabular}
    }
    \caption{ndcg@10 retrieval metric after reranking 10 documents. Learning document weights ($T=10$ steps) using gradients from Mamba2 \emph{with no finetuning}, one can typically outperform BM25 and in some cases reach close to the performance of E5-base. Asterisks (*) highlight scores on tasks which the associated retriever was directly finetuned on.}
    \label{tab:reranker_ndcg_k10}
\end{table*}

While our overall focus is on quality of the model generation, we first analyze our method using common retrieval metrics. Retrievers are traditionally evaluated in terms of how well their document scores align with a set of ground truth relevance labels \cite{muennighoff2022mteb}. If we only care about whether the retriever collects the “correct” documents within some top-$k$ retrieved, typical precision and recall metrics can be calculated. Additionally if we care about the ranking within that set, metrics such as the normalized discounted cumulative gain (nDCG) are the standard (see Appendix \ref{sec:eval_metrics} for details). All retrieval metrics presented here are normalized within $[0,1]$, with higher being better.

When evaluating our method on standard retrieval tasks, we will use the learned weights $\hat{\alpha}$ directly as the ranking score. 
We first demonstrate that \methodname document rankings produced in this manner are reasonably correlated with these ground truth relevance scores. 
In \cref{fig:hotpot_curves}, we compare the three different versions of our method to baselines specifically on the HotpotQA dataset: using multiple steps of a gradient descent procedure (mamba2 multistep grad), the single gradient retrieval conditioned on the mean document state (mamba2 grad), and lastly using the gradient without conditioning on any documents (mamba2 grad zero state). 
Baselines include BM25 \citep{roberston2009probabilistic} and two common semantic retrievers (e5-base-v2, bge-m3), which work by measuring the cosine similarity between query and document embeddings \cite{wang2022text,chen2024bge}. 
We additionally generate scores using our loss function separately for each document (mamba 2 loss paragon), as well as by naively taking the inner product between document and question states (naive mamba2).
Interestingly, we see that out of the box \methodname works better than BM25 for the HotpotQA task and is close to BGE-M3. While the multistep optimization method performs best, the single gradient methods see only about a 5\% difference in performance. 

We expand this analysis to look at a larger range of documents across different datasets and model sizes. In \cref{tab:reranker_ndcg_k10}, we can observe again that \methodname typically outperforms BM25 across retrieval tasks, and on MuSiQue and 2Wiki tasks approaches the performance of E5. Interestingly, retriever metrics seem to decline slightly but consistently for larger models ($\sim2$\% from 130m to 2.7b size models). One reason for this may be because larger models ‘knows more’ a priori (as evidenced by higher default answer accuracy) and therefore may select for certain documents less strongly. 

We stress that although this retrieval is not state of the art when measured according to the ground-truth `relevant' documents, the primary goal of RAG is to improve the quality of the final generated answer. We will see in the following experiment that \methodname can produce competitive results in this regard. 

\subsection{Generation Metrics}
\begin{table*}[h]
    \centering
    \resizebox{\textwidth}{!}{
\begin{tabular}{ll|rrrrr}
\toprule
 Model & Generation Method & MS MARCO (↑)& HotpotQA (↑)& MuSiQue (↑)& 2WikiMultihopQA (↑) & Average\\
 \midrule
  \multirow{5}{*}{Mamba2-130m}  &  Ground Truth Concatenation & 46.2 & 26.7 &  19.4 &  21.0 & 28.3\\
  \cmidrule(lr){2-7}
&  BM25 Reranking  & 48.0 & \bfseries 29.8 & 10.2 & 14.2 & 25.6\\
&  BGE-M3 Reranking  & 42.5* & 26.4* & 13.9 & \bfseries 23.0 &  26.5*\\
&  E5 Reranking & \bfseries 55.5* & 25.1 & 14.8 &  14.7 & 27.1*\\
&  Learned Reranking (Ours)  & 53.9 & 24.9 & 12.7 &  20.9 & \bfseries 28.1\\
&  Learned State (Ours)  & 45.8 & 24.5 &\bfseries 15.6 & 18.6 & 26.1\\
 \midrule
  \multirow{5}{*}{Mamba2-1.3b}  &  Ground Truth Concatenation  & 61.4 & 33.5 &  24.6 &  32.7  & 38.0\\
  \cmidrule(lr){2-7}
&  BM25 Reranking  & 60.8 & 33.7 & 13.7 & 19.7  & 32.0\\
&  BGE-M3 Reranking  &\bfseries 64.7* & 32.1* & 17.4 & 23.3 &  34.4*\\
&  E5 Reranking  & 63.2* & 29.2 & 17.3 & 26.8  & 34.1*\\
&  Learned Reranking (Ours)  &   63.7 & \bfseries 37.0 & \bfseries 23.3 & \bfseries 30.4& \bfseries 38.6 \\
&  Learned State (Ours)  & 62.4 & 30.8 & 21.7 & 19.7 & 33.6\\
\midrule
  \multirow{5}{*}{Mamba2-2.7b}  &  Ground Truth Concatenation &  72.3 & 39.8 &  29.6 &  31.5 & 43.3\\
  \cmidrule(lr){2-7}
&  BM25 Reranking & 65.2 & 37.4 & 20.7 & 23.0 & 36.6\\
&  BGE-M3 Reranking  & 69.2* & \bfseries 40.6*& 22.3 &26.0 & \bfseries 39.5*\\
&  E5 Reranking & \bfseries 70.8* & 36.7 & 21.1 & 21.7 & 37.3*\\
&  Learned Reranking (Ours) & 64.7 &  39.3 & \bfseries 23.6 & \bfseries 28.7 &  39.1 \\
&  Learned State (Ours)  & 62.9 & 37.4 & 19.9 & 18.0 & 34.5\\

\bottomrule
\end{tabular}
}
    \caption{Generation metrics (F1 score from Claude3-Haiku using Ragchecker) for different Mamba2 models after conditioning on retrieved documents. Ground truth concatenation provides the model with all documents determined relevant by the dataset and serves as a paragon. Asterisks (*) highlight scores on tasks which the associated retriever was directly finetuned on.}
    \label{tab:gen_metrics}
\end{table*}

In order to measure the quality of generated answers, we make use of the recently released evaluation pipeline Ragchecker, which uses an external model to evaluate the precision and recall of generated claims versus the ground truth answer \cite{ru2024ragchecker}. We use their overall F1 score metric, which has been shown to correlate well with human evaluation.

For these experiments we take the ground truth set of documents provided by each dataset with attached relevance scores. As a paragon, we sort documents based on this ground truth relevance and feed into the context so that the most relevant document is closest to the question (i.e. prompt=$(d_k,\dots,d_1,q)$) (this produces higher quality answers compared to the reverse sort, as shown in \cref{sec:doc_reordering}). Note that in the case of ties (i.e. multiple documents with the same relevance score) we keep their default ordering. We then evaluate different concatenations of reranked documents from BM25,BGE-M3, E5, and our method (learned reranking). We also experiment with generating from the learned state, $\bar{h}(\alpha) = \sum_{i=1}^k\alpha_{i} h_{d_i}$, directly (learned state). We find that on average, generations from our reranking method outperform BM25 and E5, and outperform BGE-M3 on average except for the Mamba2-2.7b model (where it achieves nearly identical performance). We point out that both the E5 and BGE-M3 retrievers are \textit{directly finetuned} on MS MARCO pairs, with BGE-M3 also finteuned on HotpotQA. This training most likely explains why these retrievers generally have higher accuracy for these tasks. When evaluated only on the two ``held out'' tasks, \methodname with Mamba2-2.7b produces an average F1 score of 26.15 as opposed to the next best of 24.15 from BGE-M3.

Additionally, in certain cases (e.g. for the Mamba2-1.3b model), we find that our method produces learned rerankings better than `ground truth'. This mismatch highlights potential flaws with solely relying on traditional retriever metrics for evaluating RAG pipelines. 

While using the learned state directly typically is not as effective as concatentating documents, we see how the learned state can still be useful in long-context settings in \cref{tab:gen_metrics_long_context}: for TriviaQA dataset, the full document concatenation (consisting of whole articles such as from Wikipedia) is longer than the Mamba2 training context length, however we can observe that using our learned state mixture can improve the quality of generation dramatically. Across models we achieve 97\% relative improvement in F1 score on average for this task compared to the next best retriever.  

\begin{table*}[h]
    \centering
    \resizebox{\textwidth}{!}{
\begin{tabular}{l|rrr}
\toprule
 Generation Method & TriviaQA Mamba2-130m(↑) & TriviaQA Mamba2-1.3b (↑)& TriviaQA Mamba2-2.7b (↑)\\
 \midrule
  Ground Truth Concatenation & 44.4 & 29.8 & 32.4 \\
  \cmidrule(lr){2-4}
  BM25 Reranking & 45.1 & 25.2 & 10.3\\
  E5 Reranking & 42.8 & 29.1 & 35.0 \\
 Learned Reranking (Ours) & 43.3 & 25.7 & 33.1\\
 Learned State (Ours) & \bfseries 53.6 & \bfseries 79.7 & \bfseries 69.5 \\
\bottomrule
\end{tabular}
}
    \caption{Generation metrics (F1 score from Claude3-Haiku using Ragchecker) for different Mamba2 models on a long context reranking task}
    \label{tab:gen_metrics_long_context}
\end{table*}

\subsubsection{Generations with Oracle Loss}
While conditioning generation directly on the learned state $\bar{h}(\alpha)$ does not outperform the learned reranking for most tasks evaluated, we see in \cref{tab:answer_paragon_2.7b} that we can consistently exceed performance when training on the oracle (answer) loss. This suggests that
further improvements can be made to the current loss function proxy (see \cref{sec:discussion} for discussion of these possibilities).
\begin{table*}[hbt!]
    \centering
    \resizebox{0.9\textwidth}{!}{
\begin{tabular}{lrrrrr}
\toprule
 & TriviaQA (↓)& MS MARCO (↓)& HotpotQA (↓)& MuSiQue (↓)& 2WikiMultihopQA (↓)\\
\midrule
Default & 3.538 & 1.858 & 2.549 & 3.022 & 1.710 \\
\midrule
BM25 Top 20 & 5.998 & 0.962 & 2.055 & 2.296 & 1.250 \\
Learned Reranking & 11.604 & 0.928 & 1.971 & 1.770 & 0.956 \\
Learned State & \bfseries 5.528 & \bfseries 0.865 &\bfseries 1.945 & \bfseries 1.362 & \bfseries 0.780 \\
\midrule
E5-base Top 20 & 4.429 & 0.964 & 2.142 & 2.056 & 1.210 \\
Learned Reranking & 4.241 & 0.953 & 2.016 & 1.729 & 1.041 \\
Learned State & \bfseries 2.698 & \bfseries 0.862 & \bfseries 1.861 & \bfseries 1.276 & \bfseries 0.912 \\
\bottomrule
\end{tabular}
}
    \caption{Mamba2-2.7b answer loss (cross entropy) after conditioning on different combinations of documents learned using the oracle loss function $P_\mathcal{M}(a,q|\cdot)$.}
    \label{tab:answer_paragon_2.7b}
\end{table*}


\section{Discussion}
\label{sec:discussion}
We have developed a method for incorporating retrieved documents into SSMs that can improve generation across different document and model sizes. As we are simply using the emergent properties of model gradients with respect to the input state, this method can be applied to any SSM out of the box with relatively few adjustments. Furthermore, framing document retrieval as an inference time optimization procedure opens up interesting possibilities for varying test-time compute by dynamically adjusting the number of document learning steps. 

\paragraph{Current Limitations}
While this method is effective at scoring smaller sets of documents, the memory cost of indexing the full SSM state currently prevents us from indexing document stores larger than a few thousand documents. We have explored some initial methods of compressing this state (see \cref{sec:warm_start_compression}), and found that sampling a subset of SSM layers can preserve the majority of performance, however further research is needed to determine whether this can be compressed further.

Additionally there are times when the question loss does not align with the true answer loss. For example on questions asking about multiple people, the loss is sometimes sensitive to the order in which they are included (see \cref{sec:additional_experiments} for specific examples). Generating multiple rephrasings of the same question, training on initial candidate answers, or using token-level adjustments (similar to  \cite{duan2023shifting}) may be an interesting line of future work for improving the quality of our proxy loss. Using a (differentiable) reward model in place of simpler token-loss function may be another interesting alternative.

\paragraph{Extension to Transformers} Using the state space dual (SSD) interpretation, we can show that learning a mixture of SSM states is equivalent to learning a mixture of KV caches for linear transformers. While this connection is not strictly applicable for softmax attention, we have some initial experiments showing that learning document weights in the form of the causal mask can produce intuitive relevance scores (see \cref{sec:transformers}).

\paragraph{Measuring the Value of Document Collections} Task2vec \citep{achille2019task2vec} uses the Fisher information matrix of model weights to measure the difficulty of different tasks. As we are also taking gradients of the log token loss (cross entropy of the token sequence) w.r.t. the documents in our data store, one might similarly use the this gradient norm after test-time training as a way to measure how much information the documents contain about the question. Intuitively, wide local minima indicate the model is not sensitive to the context, while narrow local minima indicate the model is heavily relying on the retrieved documents for its performance.

\section*{Impact Statement}
This paper presents work whose goal is to make language models more reliable and reduce hallucinations. There are many potential societal consequences 
of our work, none which we feel must be specifically highlighted here.
 
\newpage
\bibliography{aux/ref}
\bibliographystyle{aux/icml2024}

\newpage
\appendix
\onecolumn
\section{Extension to Transformers}
\label{sec:transformers}
The vanilla causal attention layer for the transformer architecture \citep{vaswani2017attention} is defined as a map from for the query, key and value matrices $Q,K \in\mathbb{R}^{(t,n)}, K\in\mathbb{R}^{(t,p)}$ to output sequence $Y\in\mathbb{R}^{(t,p)}$:
\begin{align}
    Y =MV = \text{softmax}(L\circ(QK^\intercal))V.
\end{align}
Here the sequence transformation matrix $M\in \mathbb{R}^{(t,t)}$ is made autoregressive by masking the query, key product with a lower triangular matrix of 1's ($L$).

As explored in previous works, the Mamba2 architecture can be interpreted as linear attention with a \emph{learned} causal mask \cite{yang2023gated,dao2024transformers, bick2024transformers}.

\begin{align}
\begin{cases}
    h_{t} = \A_t h_{t-1} + \B_t x_t \nonumber \\
    y_t = \C_t h_t
\end{cases} \quad \Leftrightarrow \quad 
    Y = MX = L_{\alpha} \circ (C\cdot B^\intercal)\cdot X
\end{align}
Where $L_\alpha$, $C$, and $B$ is defined as
\begin{align}
    L_\alpha := \begin{bmatrix}
        1 & 0 & 0 & \dots & 0\\
        \alpha_{2} & 1 & 0 & \dots & 0 \\
        \alpha_{3:2} & \alpha_{3} & 1&\dots&0\\
        \vdots &\vdots&\vdots&\ddots &\vdots \\
        \alpha_{t:2} & \alpha_{t:3}&\alpha_{t:4}&\dots&1
    \end{bmatrix} 
    \quad C:= \begin{bmatrix}
        C_1 \\C_2 \\ \vdots \\C_t
    \end{bmatrix}
    \quad B:= \begin{bmatrix}
        B_1 \\B_2 \\ \vdots \\B_t
    \end{bmatrix}
\end{align}

Note that $C$ and $B$ block matrices take the place of query and key matrices, while the scalars $\alpha_i$ contribute to the causal mask. Through this perspective, \methodname can be interpreted as optimizing different \emph{blocks} of this attention mask $L_\alpha$ at inference time.  

\begin{figure}[h]
    \centering
    \includegraphics[width=0.48\linewidth]{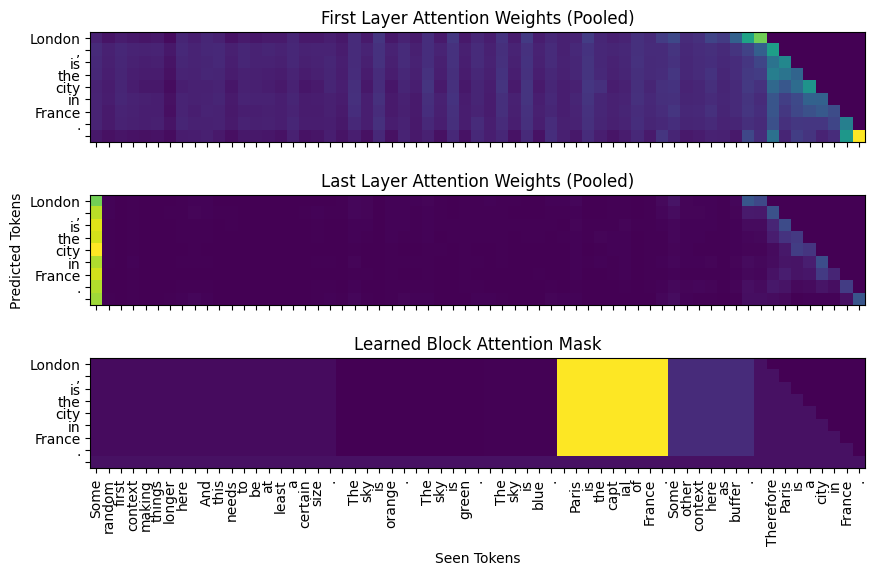}
    \includegraphics[width=0.48\linewidth]{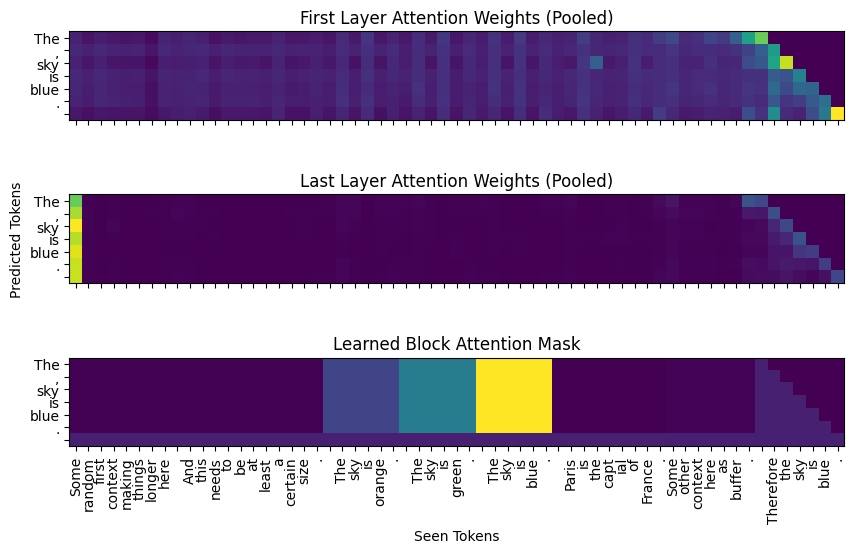}
    \caption{Simple example of "learning" attention mask blocks for the Phi-3.5-instruct transformer model. Each sentence of the context is considered a separate `document', and we attempt to learn the attention mask which maximizes the model's likelihood of outputing "Therefore Paris is a city in France" (left) and "Therefore the sky is blue." (right). Intuitively, the sentences "Paris is the captial of France" and "The sky is blue" respectively become the highest weighted blocks of the mask. In contrast, using the default attention weights does not reveal much information about which sentences are the most useful for prediction.}
    \label{fig:phi_learned_mask}
\end{figure}

While most transformer architectures use the softmax non-linearity, preventing a direct application of the \methodname framework, we run a basic experiment demonstrating that learning attention mask weights can still predict useful contexts. In \cref{fig:phi_learned_mask}, we learn attention mask weights for six context sentences using the Phi-3.5-instruct transformer model. We see that in order to minimize the loss of the final sentence, the model learns to weight the most related sentence more strongly. These results suggest interesting future investigation into whether test-time context optimization can be efficiently applied to transformers as well.

\section{Experimental Details}
Main experiments were run on an Amazon EC2 P4 instance with 8 NVIDIA A100 Tensor Core GPUs. By default \methodname uses $T=10$ steps of AdamW optimization with learning rate $\eta=1e-1$.
All answers are generated using a fixed temperature of $T=0.7$ and maximum sequence length of $500$ tokens. 

\subsection{Evaluation Metrics}
\label{sec:eval_metrics}
For computing traditional retriever metrics we use the ranx python library \cite{ranx}. In addition to precision and recall, we use the normalized discounted cumulative gain (nDCG) and mean average precision (MAP) which are commonly used to evaluate \emph{relative orderings} of top returned documents as well. 

The discounted cummulative gain for $k$ retrieved documents is calculated as \cite{ndcg}:
\begin{align}
    DCG := \sum_{i=1}^k\frac{2^{rel_i}}{\log_2(i+1)},
\end{align}
Where $rel_i$ is the ground-truth relevance score of the top $i$-th document returned. As relevant documents are returned at later indexes $i$ the denominator grows and the score is naturally diminished. Dividing by the ideal maximum value of this score (all relevant documents are returned first) produces the normalized score nDCG$\in[0,1]$.

For $R$ ground-truth relevant documents, MAP for the top-$k$ retrieval is defined as
\begin{align}
    MAP:= \frac{\sum_r Precision@r}{R},
\end{align}
where $r$ is the index of each relevant document that was returned.
For example, consider a retriever that returns three documents with ground truth relevance scores $rel_1=1,\:rel_2=0,rel_3=1$. Then $R=2$ and the MAP would be calulated as $(P@1+P@2)/2 = (1+2/3)/2 = 5/6$. 

For our generation-side metrics we use the F1 scores produced by the Ragchecker evaluation framework \cite{ru2024ragchecker}, which uses an external LLM to determine whether claims made in the prediction are entailed by the ground-truth answer. As the external model we use Claude3-Haiku, called through AWS bedrock API.

\subsection{Datasets}
\label{sec:dataset_details}

In \cref{tab:dataset_statistics} we provide the average number of documents and characters for subsets used for retrieval metric evaluations. Each dataset subset consists of 1k samples from the validation split.

\begin{table}[h]
    \centering
\begin{tabular}{lrrrrr}
\toprule
 & TriviaQA & MS MARCO & HotpotQA & MuSiQue & 2WikiMultihopQA \\
\midrule
Avg. \# Documents & 4.8 & 9.9 & 11.2 & 20.0 & 10.0 \\
Avg. \# Characters & 16668.8 & 309.7 & 511.3 & 520.7 & 368.2 \\
\bottomrule
\end{tabular}
    \caption{Statistics on the document retrieval and reranking datasets used.}
    \label{tab:dataset_statistics}
\end{table}
\section{Additional Experiments}
\label{sec:additional_experiments}
\subsection{Computation Times}
\Cref{tbl:computation} compares computational complexity of different retrieval and reranking methods. We can see that our method scales only in the number of optimization steps $T$, and not with the document store size. This means for example that a cross-encoder would need 100 model calls to rank 100 documents, while our method only needs a few forward and backward passes through the model (e.g. $T=10$). When compared to an LLM-based reranking method (e.g. directly prompting the model “please score these documents”), RICO avoids the expense of processing long contexts at inference time, as documents states have already been precomputed.

In \cref{tab:gradient_times}, we compare average gradient computation times for each of the Mamba2 models used to retrieval times of the baseline models, and see the former is 1-2 orders of magnitude slower. This motivates our use of the method primarily as a reranker, since in this setting we initialize weights close to the solution and require only a few steps of gradient descent. 

\begin{table}[h]
\centering
\resizebox{0.9\textwidth}{!}{%
\begin{tabular}{@{}lllll@{}}
\toprule
\textbf{Retrieval Generation Method}                               & \textbf{Model Calls @ Inference} & \textbf{Context Size per Call} & \textbf{Additional Computation} & \textbf{Examples}       \\ \midrule
Bag-of-words                                                       & –                                & –                              & word frequency counts           & BM25                    \\
Dense Embedding                                                    & 1                                & query length                   & N dot products                  & E5, BGE-M3              \\
Context Optimization (Ours)                                        & 2T                               & fixed state size               & N*T dot products                & RICO (ours)             \\
Cross-Encoder                                                      & N                                & doc length + query length      & –                               & SGPT                    \\
\begin{tabular}[c]{@{}l@{}}LLM Reranking (batched) \\ \end{tabular} & N/B                              & B * doc length + query length  & –                               & RankGPT, LLM Likelihood, SeaKR \\ \bottomrule
\end{tabular}%
}
\caption{Computational comparison of different retrieval+generation methods in terms of model calls at inference time. Here N is the number of candidate documents to rerank, T is the number of optimization steps for RICO (typically a small number [1-10]) , and B is a batch size parameter (B=1 is the case where each document is scored individually by the LLM). }
\label{tbl:computation}
\end{table}

\begin{table}[h]
    \centering
\begin{tabular}{@{}lr@{}}
\toprule
Model       & Gradient Time (ms) \\ \midrule
Mamba2-130m & 192                            \\
Mamba2-1.3b & 388                            \\
Mamba2-2.7b & 522                            \\ \bottomrule
\end{tabular}
\quad\quad
\begin{tabular}{@{}lr@{}}
\toprule
Method      & \multicolumn{1}{l}{Retrieval Time (ms)} \\ \midrule
bm25       & 2                                       \\
e5-base-v2 & 8                                       \\
bge-m3     & 39                                      \\ \bottomrule
\end{tabular}
    \caption{(Left) Average time to compute gradients w.r.t. the input state (forward+backward pass through the model) across benchmarks for Mamba2 models. (Right) Average time of retrieval over benchmarks using document stores of size 1-2k.}
    \label{tab:gradient_times}
\end{table}

\subsection{State Compression and Warm Start Retrieval}
\label{sec:warm_start_compression}
\begin{figure}
     \centering
     \includegraphics[width=0.5\linewidth]{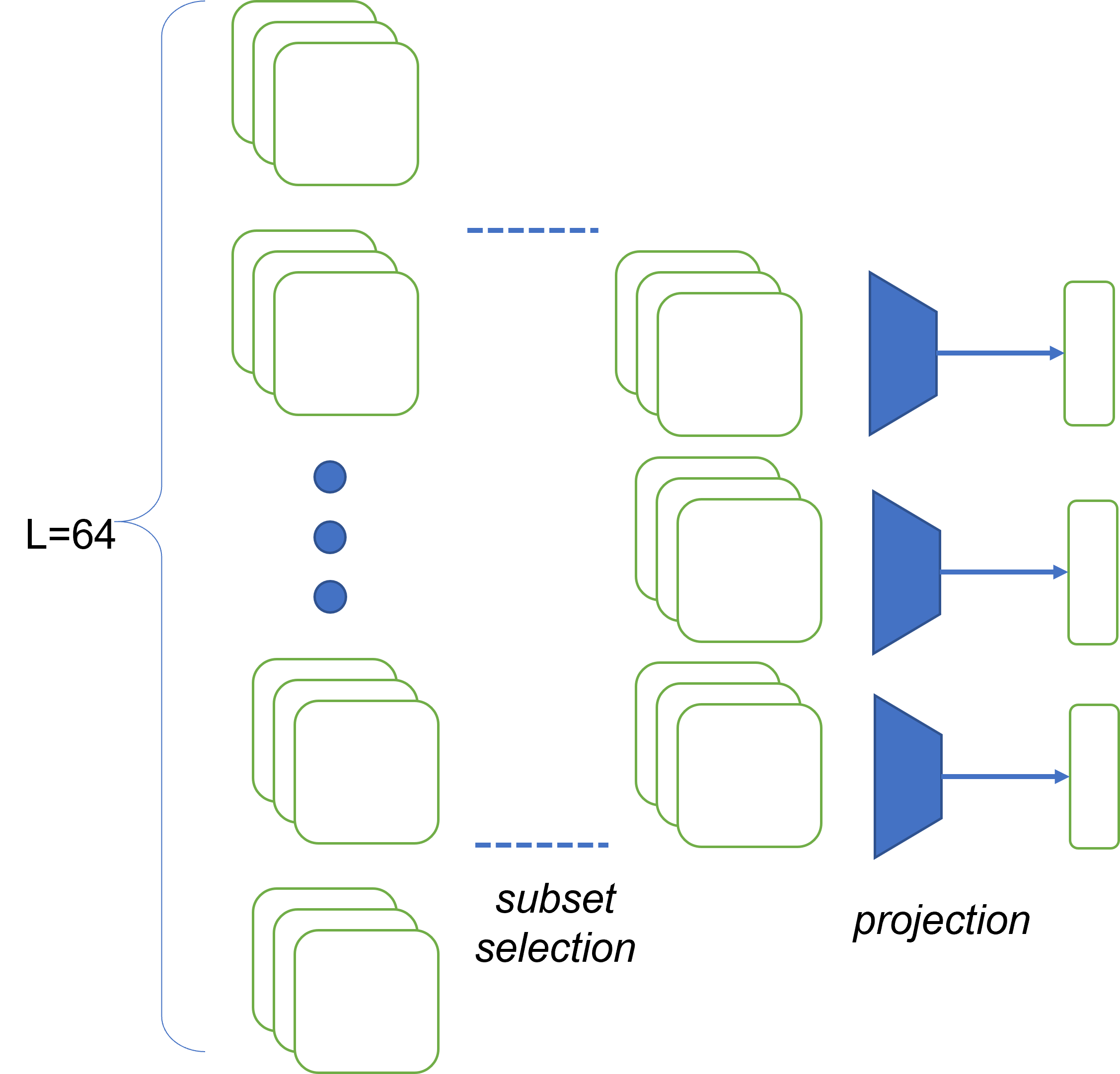}
     \caption{Diagram of state subsampling and compression method for Mamba2-2.7b model with 64 layers.}
     \label{fig:compression diagram}
 \end{figure}

Due to the large size of full Mamba2 internal states (even the Mamba2-130m model for example contains on the order of $\sim 1$M parameters), we explored some simple compression methods including layer-wise random projection and PCA (see \cref{fig:compression diagram}). While we found that these methods can decrease the size of the state embeddings used, the best accuracy vs. compression tradeoff was achieved by simply taking a subset of middle layers from each model (4/24, 8/48, 10/64 layers for 130m, 1.3b, and 2.7b models respectively). In \cref{tab:topk_ndcg_k10}, we explore using a combination of layer subsampling and warm start initialization (using BM25 as a coarse retriever) to perform top-$k$ retrieval over document stores 1-2k in size (consisting of documents associated with 1k questions from each dataset sample). We find that on average this variation of the method outperforms BM25 and BGE-M3 on the nDCG@10 retrieval metric, while underperforming the E5 retriever. 

\begin{table*}[h]
    \centering
    \resizebox{0.9\textwidth}{!}{
    \begin{tabular}{llrrrrr}
    \toprule
     \bfseries Retriever&\bfseries Method & MS MARCO (↑) & HotpotQA (↑) & MuSiQue (↑)& 2WikiMultihopQA (↑) & Average\\
    \midrule
    BM25&---& 0.55 & 0.745 & 0.435 & 0.664 &0.599\\
    BGE-M3 &---& 0.638&	0.760&	0.502&	0.642&0.636 \\
    E5-base &---& 0.711 & 0.792 & 0.621 & 0.78 &0.726\\
    \cmidrule{2-7}
    \multirow{3}{*}{Mamba2-130m (layer subset)} & Gradient &0.585&	0.678&	0.445&	0.574& 0.571\\
    & Multistep Gradient & 0.598&	0.735&	0.478&	0.669&0.620\\
    & WS Multistep Gradient &0.572&	0.746&	0.578&	0.743&0.660\\
    \cmidrule{2-7}
    \multirow{3}{*}{Mamba2-1.3b (layer subset)} & Gradient & 0.555& 0.689 & 0.468&0.611&0.581 \\
    & Multistep Gradient & 0.573 & 0.750 & 0.539 & 0.700&0.641\\
    & WS Multistep Gradient &0.529 & 0.742 & 0.618 & 0.744&0.658\\
    \cmidrule{2-7}
   \multirow{3}{*}{Mamba2-2.7b (layer subset)} & Gradient & 0.554&	0.621&	0.403&	0.486&0.516\\
    & Multistep Gradient & 0.546&	0.696&	0.438&	0.574&0.564\\
    & WS Multistep Gradient &0.551&	0.732&	0.577&	0.729&0.647\\
    \bottomrule
    \end{tabular}
    }
    \caption{ndcg@10 retrieval metric after retrieving the top 10 documents from a small datastore (1-2k documents for each dataset). }
    \label{tab:topk_ndcg_k10}
\end{table*}

\subsection{Analysis of the Question Loss Landscape}
\label{sec:loss_landscape}
\begin{figure}
    \centering
    \includegraphics[width=0.4\linewidth]{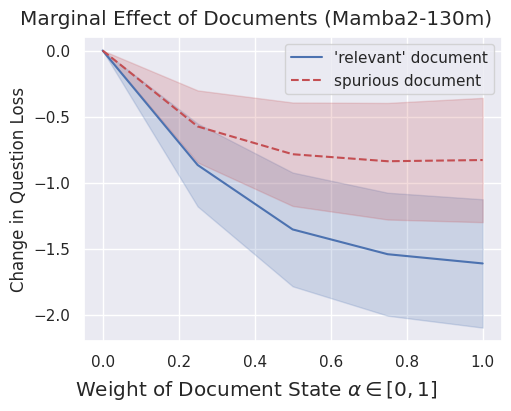}
    \caption{Plot illustrating the average change in question likelihood for Mamba2-130m model evaluated on 1k HotpotQA questions as supporting documents are added (via increase in the weight parameter $\alpha_i$). Bolded blue and red lines denote the mean loss at each weight level, while shaded area shows the standard deviation of each distribution.}
    \label{fig:marginal_loss}
\end{figure}

In \cref{fig:marginal_loss}, we can observe two phenomena: first is that marginal loss landscapes appear to be approximately convex as we change single document weights $\alpha_i$. Secondly, the cross entropy loss of the question is lower when conditioning on relevant documents (as determined by the dataset) as opposed to spurious ones. Interestingly, even these spurious documents tend to increase question likelihood slightly, most likely because they are still related to the subject matter in the question. 

\begin{figure}[h]
    \centering
    \includegraphics[width=0.65\linewidth]{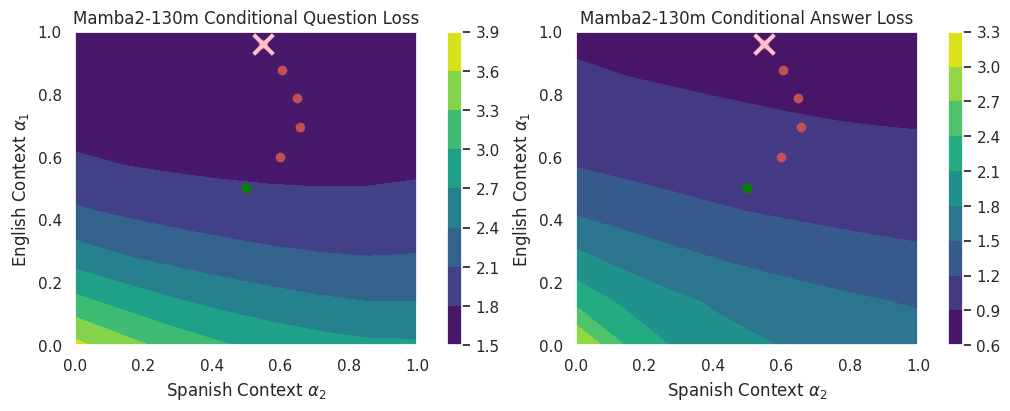}
    \caption{Token cross-entropy loss for question (left) and ground-truth answer (right) conditioned on combinations of Spanish and English context. Red dots represent a gradient optimization trajectory over the question loss landscape starting from the mean state (green).}
    \label{fig:language_example}
\end{figure}

\paragraph{Motivating Example: Incompatible Languages}
Mamba2 was trained on the Pile, an English-only dataset \cite{dao2024transformers}. If the retriever used for RAG is multi-lingual and retrieves non-English documents, it will not be useful as the generating model cannot make use of the information. For example the multilingual BGE-M3 scores both English and Spanish versions of `An orchestra conductor works in a concert hall' as similar relevance ($score_1=0.86$,$score_2=0.81$) for the question $q=$`Where does a conductor work?'. However, our model-aware method correctly places much more emphasis on the English version of the context ($score_1=0.97$, $score_2=0.53$), successfully minimizing the ground-truth answer loss in \cref{fig:language_example}.

\paragraph{Motivating Example: Avoiding Redundant Information}
Exisiting retrievers will always return the same documents regardless of what the generating model has already seen. In the following example we provide a query that already contains some useful information $q=$`Mary is 30. Question: Who is older, Mary or Jane?'. When given a useful context $d_1=$`Jane is 42' and redundant context $d_2=$`Mary is 30', a retriever like BGE-M3 will score both contexts as similar relevance when provided just the question part of the query ($score_1=0.76$,$score_2=0.75$), and even prefer the redundant information if we embed the entire query ($score_1=0.75$,$score_2=0.89$). In contrast, \methodname correctly places more emphasis on the new context ($score_1=0.96$,$score_2=0.28$), as this ranking is \emph{conditioned} on the information already provided to the model in the query.

\begin{figure}[h]
    \centering
    \includegraphics[width=0.65\linewidth]{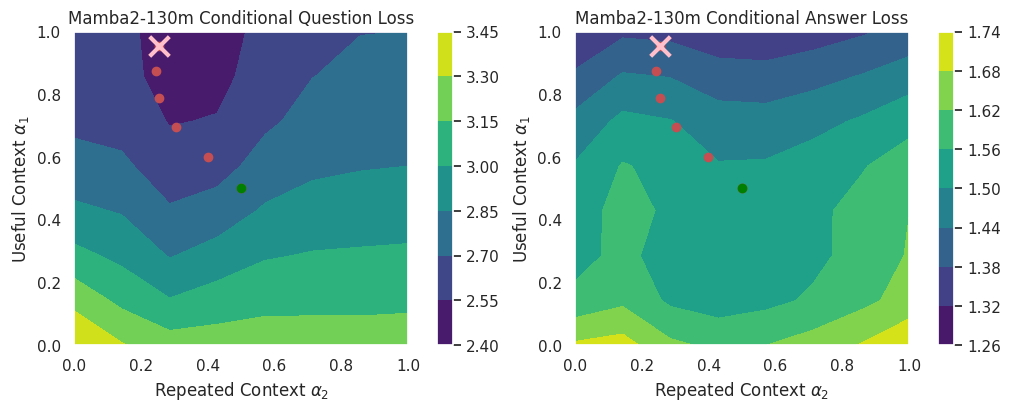}
    \caption{Token cross entropy loss for question (left) and ground-truth answer (right) conditioned on combinations of two contexts. The red dots represent a gradient optimization trajectory over the question loss landscape starting from the mean context state (green).}
    \label{fig:redundant_example}
\end{figure}

\begin{figure}[h]
    \centering
    \includegraphics[width=0.65\linewidth]{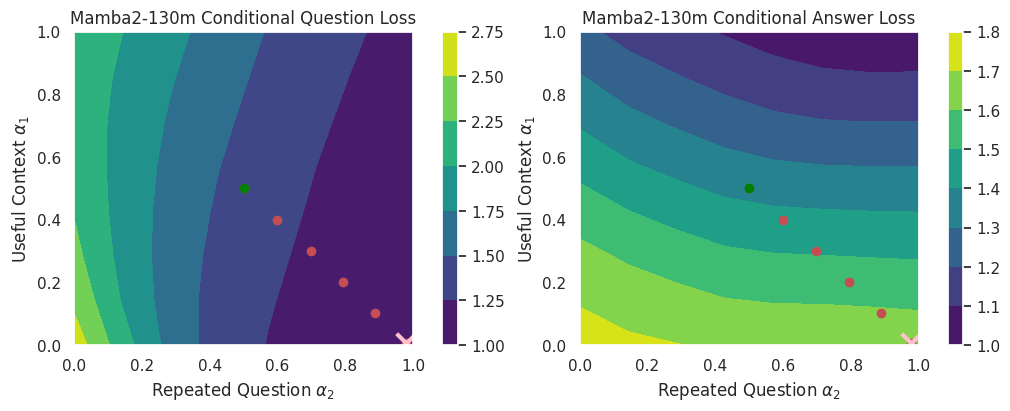}
    \caption{Example failure case of \methodname in which the question loss (left) and ground-truth answer loss (right) are not well aligned. Token cross-entropy loss landscapes are shown for different mixtures of two contexts: $\alpha_1$ associated with the useful fact ad $\alpha_2$ associated with the repeated question. Red dots repersent a gradient optimization trajectory over the quetion loss landscape starting from the mean state (green).}
    \label{fig:failure_case}
\end{figure}

\paragraph{Example Failure Case}
In \cref{fig:failure_case}, we can observe a case where the proxy and oracle loss objectives are not well aligned due to the tendency for Mamba2 models to `repeat themselves'. Here the query is $q=$`In what year was President Obama born?', along two contexts consisting of the relevant fact $d_1=$`President Obama was born in 1961' and the repeated question $d_2=q$. We see that while the question loss is minimized when the Mamba2-130m model sees the question repeated in the context, while the ground-truth answer loss is minimized by seeing the relevant fact. Token-level adjustments to the loss such as those used in \cite{su2024dragin}, or minimizing the loss of initial candidate answers may be able to alleviate such issues, and is an interesting line of future work.
\FloatBarrier

\subsection{Additional Document Ordering Experiments}
\label{sec:doc_reordering}
For the following experiment, we compare the effects of reordering the most relevant documents before the question. In \cref{tab:default_ordering}, we see that sorting the most relevant documents to appear immediately before the question (last) increases the likelihood of the correct answer when compared to placing the most relevant document first. This matches intuition for state space models, since memory `fades' overtime as the gap becomes larger between document and question.

\begin{table*}[h]
    \centering
    \resizebox{\textwidth}{!}{
\begin{tabular}{ll|rrrr}
\toprule
 Model & Generation Method & MS MARCO (↓)& HotpotQA (↓)& MuSiQue (↓)& 2WikiMultihopQA (↓) \\
 \midrule
  \multirow{2}{*}{Mamba2-130m}  &  Most Relevant Last & \bfseries -1.206&\bfseries	-0.703&\bfseries	-1.199&\bfseries	-0.795 \\
&  Most Relevant First  & -1.131&	-0.511&	-0.433&	-0.662 \\
 \midrule
  \multirow{2}{*}{Mamba2-1.3b}  &  Most Relevant Last  &\bfseries -0.916&\bfseries	-0.544&\bfseries	-1.054&	\bfseries -0.725 \\
&  Most Relevant First  & -0.881&	-0.356&	-0.504&	-0.576  \\
\midrule
  \multirow{2}{*}{Mamba2-2.7b}  &  Most Relevant Last  &\bfseries  -0.895&	\bfseries-0.501&\bfseries -1.189&\bfseries	-0.708 \\
&  Most Relevant First  & -0.852&	-0.290&	-0.331&-0.527\\

\bottomrule
\end{tabular}
}
    \caption{Change in answer loss after conditioning on documents for different orderings of the ground truth set}
    \label{tab:default_ordering}
\end{table*}

\end{document}